\documentclass[10pt, a4paper]{article}
\usepackage{lrec}
\usepackage{hyperref}
\usepackage{latexsym}
\usepackage{danudefs}
\usepackage{algorithmic}
\usepackage{algorithm}
\usepackage{color}
\usepackage{booktabs}
\usepackage{xspace}
\usepackage{url}
\usepackage{cleveref}

\usepackage[utf8]{inputenc}
\usepackage{CJKutf8}
\newcommand{\jp}[1]{\begin{CJK}{UTF8}{min}#1\end{CJK}}

\usepackage{array,multirow,graphicx}
\usepackage{float}
\usepackage{color, colortbl}

\definecolor{LightCyan}{rgb}{0.88,1,1}
\definecolor{LightYellow}{rgb}{1,1,0.88}
\definecolor{LightPink}{rgb}{1,0.71,0.75}

\title{Language-Independent Tokenisation Rivals Language-Specific Tokenisation for Word Similarity Prediction}

\name{$\text{Danushka Bollegala}^{*,\dagger}$, $\text{Ryuichi Kiryo}^{\dagger}$, $\text{Kosuke Tsujino}^{\dagger}$, $\text{Haruki Yukawa}^{\dagger}$}

\address{$\text{University of Liverpool}^{*}$, $\text{Amazon}^{\dagger}$ \\ \{danubol,kiryor,kosutu,hyukawa\}@amazon.com}

\abstract{
Language-independent tokenisation (LIT) methods that do not require labelled language resources or lexicons have recently gained popularity because of their applicability in resource-poor languages. Moreover, they compactly represent a language using a fixed size vocabulary and can efficiently handle unseen or rare words. On the other hand, language-specific tokenisation (LST) methods have a long and established history, and are developed using carefully created lexicons and training resources. Unlike subtokens produced by LIT methods, LST methods produce valid morphological subwords.
Despite the contrasting trade-offs between LIT vs. LST methods, their performance on downstream NLP tasks remain unclear. 
In this paper, we empirically compare the two approaches using semantic similarity measurement as an evaluation task across a diverse set of languages.
Our experimental results covering eight languages show that LST consistently outperforms LIT when the vocabulary size is large, but LIT can produce comparable or better results than LST in many languages with comparatively smaller (i.e. less than 100K words) vocabulary sizes, encouraging the use of LIT when language-specific resources are unavailable, incomplete or a smaller model is required. 
Moreover, we find that smoothed inverse frequency (SIF) to be an accurate method to create word embeddings from subword embeddings for multilingual semantic similarity prediction tasks.
Further analysis of the nearest neighbours of tokens show that semantically and syntactically related tokens are closely embedded in subword embedding spaces.
\\ \newline  }

\begin{document}

\maketitleabstract

\section{Introduction}

% explain the problem of tokenisation 
% explain two types of tokenisations 
% describe their pros/cons 
% mention that sub-tokenisation has been evaluated only within NMT (there are some other evaluations as well)
% goal of this paper and findings

One of the first steps in many NLP pipelines is \emph{tokenisation} -- the process of splitting a given text into a sequence of continuous lexical units for the purpose of representing the given text.
Tokenisation can be performed at various granularities such as at phrase-level, word-level, sub-word level or character-level, considering the level of textual representation required for a particular task~\cite{riedl-biemann-2018-using}.
For example, for information retrieval, we must ensure that both documents and user queries are tokenised in a consistent manner considering the relevance of the search results. 
In specialised domains such as biomedical, proper tokenisation can significantly improve the retrieval accuracy by up to $80\%$~\cite{Jiang:2007}.

Exceedingly finer tokenisation is likely to return many irrelevant results with incorrect or partial matches, whereas not tokenising larger phrases will return zero results. 
In this paper, we use the term \emph{token} to refer to both words as well as subwords, which might not necessarily be morphological units but character $n$-grams.
For example, given the string ``Hello\_world'', where ``\_''  denotes the space character, a possible sequence of subtokens could be \emph{H/el/l/o/\_/world}.
As can be seen from this example, some of the subtokens such as  \emph{H}, \emph{el},, \emph{l} are not valid English words, whereas some such as \emph{world} are. 
Therefore, the effect of subtokenisation on downstream NLP tasks that require the semantics of the original input string to be retained remains unclear.

The complexity of the tokenisation problem is language dependent.
For example, punctuation rules, delimiter characters etc. have found to be adequate to tokenise non-agglutinative languages such as English or Italian~\cite{moreau-vogel-2018-multilingual}, whereas non-white space delimited languages such as Japanese or Chinese require more sophisticated methods that jointly perform Part of Speech (PoS) tagging with tokenisation~\cite{Kudo:EMNLP:2004}.
Moreover, hyphenated words, acronyms that use punctuations must be treated as single tokens in most NLP applications, which makes tokenisation a complex problem.

Tokenisation methods can be classified into language-specific tokenisation (LST) and language-independent tokenisation (LIT).
LST methods require  lexicons for the language under consideration and are often trained on manually tokenised corpora.
The accuracy of LST depends on the coverage and quality of the linguistic resources used to train them.
%Training accurate LST for resource poor languages is often difficult. 
In particular, when the coverage of the training resources are poor such as for rare words, named entities or neologisms, the accuracy of tokenisation of out of vocabulary (OOV) words can be low.
LST methods have been trained using different sequence labelling methods such as hidden Markov models (HMMs)~\cite{Jurish:2013}, conditional random fields (CRFs)~\cite{Kudo:EMNLP:2004} and recurrent neural networks (RNNs)~\cite{morita-etal-2015-morphological}.

LIT has gained popularity as an alternative to LST~\cite{Sennrich:2016,zhu-etal-naacl2019,kudo-richardson-2018-sentencepiece,unigram:kudo,Schuster_2012} because, unlike LST, LIT methods do not require predefined vocabularies nor manually tokenised texts, and operate on statistical information obtained from a large text corpora. 
For example, text compression methods such as byte pair encoding (BPE)~\cite{Gage:1994,Sennrich:2016} and language modelling (LM) methods~\cite{unigram:kudo} automatically select frequent subwords as tokens, and segment a given text such that some loss function (e.g. negative likelihood or code length) is minimised.
LIT has become the de-facto standard in text generation applications such as Neural Machine Translation (NMT)~\cite{W18-1810}, where a small vocabulary size is preferred for speeding up the decoding process~\cite{Bahdanau:2014}.
Moreover, subword regularisation using probabilities produced by the LM approach has shown to improve the accuracy of NMT~\cite{unigram:kudo}.
However, as seen from our previous example, unlike LST, LIT often produces nonsensical subwords, which are not valid morphological units~\cite{zhu-etal-naacl2019}.

As discussed above, LST and LIT have complementary trade-offs.
It remains unclear whether the loss in morphological information and the noise introduced by LIT out weights the benefits of using a small and fixed vocabulary, enabling us to overcome OOV issues across typologically diverse languages.
To empirically answer this question, we compare LST vs. LIT for multilingual lexical semantic similarity prediction for the eight languages: English (\textbf{en}), German (\textbf{de}), Spanish (\textbf{es}), Farsi (\textbf{fa}), Italian (\textbf{it}), Japanese (\textbf{ja}), Turkish (\textbf{tr}) and Thai (\textbf{th}).
Our contributions and findings in this paper can be summarised as follows:
\begin{itemize}
\item We independently conduct LST and LIT on eight languages and use Global Vectors (GloVe)~\cite{Pennington:EMNLP:2014} to learn word embeddings. 
We then predict the semantic similarity between two words using the learnt word embeddings, and measure the correlation against human similarity ratings across a suite of benchmark datasets. 
\item We evaluate different methods to compose word embeddings from subword embeddings and find that Smoothed Inverse Frequency (SIF) \cite{Arora:ICLR:2017} to outperform simple averaging, which has shown to be a strong baseline in prior work.
\item For LIT methods, for the first time, we compare BPE and LM in terms of both tokenisation speed and their accuracies for predicting semantic similarity between words.
\item Our experimental results show that for smaller (less than 100K tokens) vocabularies, LIT consistently outperforms LST. 
Moreover, between LIT methods, LM outperforms BPE. 
\end{itemize}

%We note that some work has been done on NMT and other tasks (morpholigy and BPE only). 
%We do both BPE and LM. Compare speed as well. 
%We do this for many languages and consider averaging vs. SIF for creating token embeddings from subtoken embeddings (this has not been done before and turns out to be superior!)

Our goal in this paper is \emph{not} to propose novel methods for LST or LIT. Instead, our objective is to compare LST and LIT for word embedding learning, and empirically evaluate the differences across a diverse set of languages using semantic similarity prediction as an evaluation task.
Tokenisation is one of the fundamental pre-processing steps in any NLP pipeline and has a long and established history of numerous approaches.
Although we cannot hope to conduct an extensive survey of all prior tokenisation methods due to space limitations, we briefly summarise the background details of LIT and LST methods in Section~\ref{sec:background} to support the readers to understand the experimental results described in the paper.
Several prior work have already investigated the effect of subtokenisation for different NLP tasks. 
We describe these related prior work in Section~\ref{sec:related} and highlight the important differences between the findings reported in this paper.
Evaluation protocol and experimental results comparing LST vs. LIT methods are described in Section~\ref{sec:eval}.

\section{Background}
\label{sec:background}

\subsection{Language Specific Tokenisation}

LST methods use language-specific resources such as lexicons, manually tokenised corpora and/or language-specific rules.
Earlier versions of the Stanford Core NLP toolkit~\cite{manning-etal-2014-stanford} internally used JFLex\footnote{\url{https://jflex.de/}}, a meta language for specifying tokenisation rules based on regular expressions and procedures, to execute when a rule matches.
Unlike the statistical tokenisers, rule-based tokenisers are easier to debug and their behaviour is deterministic.
For example, a product name might be required to tokenise in a specific manner, which is easier to specify as a rule rather than having to prepare numerous manually tokenised examples of contexts to train a model.
For those reasons, rule-based tokenisers have been used extensively in industrial NLP applications either as a standalone module or in conjunction with statistical tokenisers~\cite{remus-etal-2016-empirist}.

Statistical or machine learning-based tokenisation methods model tokenisation as a sequence labelling problem where we must predict whether a token boundary must be placed at a given position in an input text string. 
For example, information about the current token and its context such as previous or following tokens can be used as features for training a sequence labeller such as a hidden Markov model~\cite{papageorgiou-1994-japanese}, conditional random field~\cite{Kudo:EMNLP:2004} or a recurrent neural network~\cite{chen-etal-2015-long}.
In languages such as Japanese or Chinese where multiple possible tokenisations of the input string exist, one must find the most likely sequence of tokens~\cite{Kudo:EMNLP:2004}.
This can be modelled as a dynamic programming problem and solved efficiently via forward-backward inference methods.
Moreover, token boundaries as well as morphological properties of the tokens such as their part-of-speech (POS) tags can be simultaneously determined, which is known as morphological analysis.

To train statistical tokenisers we need lexicons, which lists all the words in a language, and manually tokenised texts as the training data.
Words that do not occur in the lexicon (i.e. out of vocabulary words) can get incorrectly tokenised and is a major cause of errors in statistical tokenisers.
Moreover, manually tokenised texts might not be available for the domain in which we might want to use the tokeniser after training, and manually creating such labelled training data can be both costly as well as time consuming.

\subsection{Language Independent Tokenisation}

Tokenising texts into \emph{subwords/subtokens}, lexical units smaller than words/tokens, has received much attention lately with their effectiveness in deep learning-based NLP models. For example, named entities, cognates/loanwords, and morphologically complex words that contain multiple morphemes are extremely challenging to properly tokenise because the occurrences of such terms are rare even in large training datasets.
On the other hand, substrings of such terms are likely to be more frequent. 
Tokenising texts into subtokens has been sufficient for a broad range of NLP tasks such as machine translation~\cite{Sennrich:2016} and language modelling~\cite{pires-etal-2019-multilingual}, where tokens are represented using lower-dimensional embedding vectors and fed into deep learning architectures.

\newcite{Sennrich:2016} proposed a subtokenisation method inspired by BPE, which is a data compression technique that iteratively replaces the most frequent pair of bytes in a sequence with a single unused byte.
Specifically, the set of symbols (i.e. symbol vocabulary) is initialised with the set of characters and each word is represented as a sequence of characters, plus a special end-of-word symbol. This is useful if we want to restore the original tokenisation after subtokenising.
Next, BPE iteratively counts all symbol pairs and replaces each occurrence of the most frequent pair (`A', `B') with a new symbol `AB'.
Each merge operation produces a new symbol, which represents a character $n$-gram.
Frequent character $n$-grams (or whole words) are eventually merged into a single symbol.
Because of this bottom-up nature of BPE, it does not require a shortlist and the final symbol vocabulary size is equal to the size of the initial vocabulary, plus the number of merge operations.
This is ideal for producing smaller vocabularies in natural language generation tasks such as machine translation to reduce the GPU memory footprints and the training time because every element in the output vocabulary is a potential candidate for generation.
The number of merge operation is a hyperparameter in BPE that can be tuned to generate arbitrarily smaller vocabulary sizes.

An alternative subtokenisation method was proposed by \newcite{unigram:kudo} based on the unigram language model under the assumption that each subword occurs independently, and consequently, the probability of a subword sequence can be computed as the product of the individual subword occurrence probabilities.
This method iteratively increases the size of the vocabulary (set of subtokens) such that a user-defined limit is reached.
It computes the optimal set of subtokens based on their occurrence probabilities, estimated using the expectation maximisation (EM) algorithm.
The initial seed vocabulary can be set to the union of all characters and the most frequent substrings in the corpus.
Because the vocabulary contains all individual characters in the corpus, subtokenisation using the unigram language model produces a probabilistic mixture of characters, subtokens and word segmentations.

Both BPE and unigram language model can be trained using untokenised text corpora. 
Moreover, both methods can be used independently of the language, which make them ideal candidates for tokenising resource poor languages.
Because of those reasons BPE and unigram language model are considered as LIT methods to compare in this paper.

\section{Related Work}
\label{sec:related}

% differentiate with other papers who have done similar evaluations.
% Vulic NAACL, Angelica and others mentioned in the first paper.
% We are not interested in morpholical segmentations because they are still language-specific. We use spaCy as a way of representing all language-specific tokenisation methods. Morpholical information might not be available for some languages and they also require morphologically annotated texts for training. So we are not interested in comparing those.

Learning embeddings for the subtokens produced by LIT methods has shown to be an effective method to overcome data sparseness issues encountered when training named entity recognisers for low-resource languages such as Uyghur and Bengali~\cite{Chaudhary:2018}.
By modelling a word as a bag of subtokens and combining pre-trained subtoken embeddings to represent rare out-of-vocabulary (OOV) words, 
\newcite{Zhao:2018} obtained SoTA results for joint prediction of POS tagging and morphosyntactic attributes in 23 languages.
These prior work show that LIT can be used to overcome OOV and rare word related issues and is especially effective for resource poor languages, but did not perform a systematic comparison between LIT vs LST methods for those tasks.

\newcite{zhu-etal-naacl2019} compared supervised morphological segmentation (SMS) by CHIPMUNK, \textbf{Morfessor} (a family of generative probabilistic models for unsupervised morphological segmentation) and BPE.
They train word and subword embeddings using skip-gram with negative sampling (SGNS)~\cite{Milkov:2013}.
They used multilingual word similarity, universal dependency parsing and fine-grained entity typing as the evaluation tasks.
They found that subword SGNS embeddings outperform subword-agnostic SGNS embeddings for morphologically richer languages such as Finnish and Turkish. SMS, which is trained according to the readily available gold standard morphological segmentations, performs best for word similarity but worst for entity typing. 
Compared to BPE, which produces short and nonsensical subwords, Morfessor is a conservative segmenter that captures longer subwords.
Consequently, Morfessor reports the best performance on entity typing. 
More importantly they emphasise that there is no single configuration that outperforms the others in all three tasks, which demonstrates the challenges involved in using subword information in a consistent manner across languages and tasks.
Moreover, addition, elementwise multiplication of subword embeddings, and self-attention are used as the composition functions for creating word embeddings from subword embeddings. 
They found that addition to be an extremely robust composition function across languages and tasks.
Surprisingly, the more sophisticated self-attention reports poor performance in many tasks.
In this paper, we propose the use of smoothed inverse frequency (SIF), which was originally proposed by \newcite{Arora:ICLR:2017} for creating sentence embeddings from word embeddings, for the purpose creating word embeddings from subword embeddings.

%\newcite{unigram:kudo}
%Creates a vocabulary and learns probabilities of subwords that maximises the likelihood over the sentences. Sort the subwords by loss (negative likelihood for example), and retain top 80\%. Do not drop unigram (individual charachters to avoid OOVs). Increase the vocabulary iteratively until we reach the desired size. Viterbi algorithm is used to find the optimal (maximum likelihood) segmentation of a given sentence.
%The paper also propsoes subword regularisation, where subword sengemts are sampled randomly according to their probabilities during training an NMT system. This cannot be done with BPE because it does not produce probabilities. Subword regularisation can be seen as a dropout via noise and improves NMT.

\section{Evaluation Protocol}
\label{sec:eval}

Evaluating tokenisation methods is a challenging task because there is no universally agreed gold standard for tokenisation~\cite{habert1998towards,webster-kit-1992-tokenization}.
Tokenisation depends both on the language as well as the task for which it is used.
Although there are some manually tokenised texts such as the Penn Treebank dataset~\cite{marcus-etal-1994-penn} for English and  Kyoto University corpus~\cite{Kawahara:2002} for Japanese that can be used to train and evaluate LST methods, no such resources are available for LIT evaluation. 
Indeed, given that the subtokens produced by LIT methods are arbitrary and depends on the size of the vocabulary specified by the user and the statistics in the corpus used to train the LIT method, what is a valid LIT of a given text remains undefined in the first place.
Therefore, following prior work comparing LST and LIT methods, we resort to an extrinsic evaluation approach where we use the tokenised output produced by a particular tokenisation method to solve an NLP task and measure its performance.

To evaluate the ability of LST and LIT methods for producing semantically meaningful tokens, we first tokenise a given text corpus using a particular tokenisation method and then use a word embedding learning method to learn embeddings for the generated tokens.
Next, we use the learnt embeddings to compute the similarity between two words and compare that with the similarity ratings assigned by human annotators for those two words.
If there exists a high degree of correlation between the predicted similarity scores and the human ratings, then it follows that the tokens produced by the employed tokenisation method correctly preserves the semantic information about words.
Semantic similarity prediction has been used as an evaluation task in prior work comparing tokenisation methods~\cite{zhu-etal-naacl2019}.

To cover a diverse set of languages with different tokenisation complexities, we select English (\textbf{en}), German (\textbf{de}), Spanish (\textbf{es}), Farsi (\textbf{fa}), Italian (\textbf{it}), Japanese (\textbf{ja}), Turkish (\textbf{tr}) and Thai (\textbf{th}).
For each of those languages we downloaded the March 2019 Wikipedia dump\footnote{\url{https://dumps.wikimedia.org}} and used Wikiextractor\footnote{\url{https://github.com/attardi/wikiextractor}} to extract texts. 
We then used the Pragmatic Segmenter\footnote{\url{https://github.com/diasks2/pragmatic_segmenter}} to split each Wikipedia article into a set of sentences.

For LST, we used spaCy\footnote{\url{https://spacy.io/}} with its corresponding pre-trained LST models\footnote{\url{https://spacy.io/models}} for \emph{en}, \emph{de}, \emph{fr}, \emph{es}, \emph{it}, \emph{fa}, \emph{th} and \emph{tr}.
For \emph{ja}, we used the CRF-based Japanese tokeniser MeCab\footnote{\url{https://github.com/taku910/mecab}} with the Japanese IPA dictionary (IPAdic) as the backend of spaCy.
\autoref{tbl:corp} shows the numbers of sentences extracted and the unique tokens for each language.

For LIT, we consider BPE and unigram language modelling (LM) both implemented in sentencepiece\footnote{\url{https://github.com/google/sentencepiece}}. 
Specifically, we randomly shuffle sentences in each corpus and train LM models with vocabulary sizes of 20K, 50K, 100K and 1M tokens.
For BPE, we train models with vocabulary sizes of 20K, 50K and 100K tokens. 
As discussed later in \autoref{sec:results}, training time of BPE is significantly longer compared to that of LM, which prevented us from creating 1M model for BPE.
The character coverage rate and maximum sentence length in sentencepiece are set respectively to 1.0 and 16384 to cover 99\% of sentences in the corpora.

\begin{table}[t]
\centering
\begin{tabular}{lcc}\toprule
Language & \#sentences & \#LST tokens \\ \midrule 
en & 	98,382,467 &	2,441,459,380 \\
de & 	43,733,620 &	860,259,675 \\
es & 	21,824,361 &	616,392,562 \\
fa &	4,334,205 &	82,277,928 \\
it	& 16,888,201	& 489,437,122 \\
ja	& 19,258,206	& 547,956,927 \\
tr & 4,006,783 & 62,444,210\\
th & 777,397 & 40,105,530\\
\bottomrule
\end{tabular}
\caption{Sizes of corpora used in the experiments}
\label{tbl:corp}
\end{table}

\subsection{Token Embedding}
\label{sec:embed}

% Explain Glove embedding and three compositional methods. Cite more recent work as well and reconfirm addition.

For corpora tokenised by LST and LIT methods, we use GloVe to learn separate token embedding sets for each language.
We set the co-occurrence window size to 15 tokens and the frequency threshold ($x_{\max}$) to 100 in our experiments.
We trained 100 and 300 dimensional token embeddings and found the latter to perform better in our experiments across languages.
Due to the space limitations, we show experimental results only for 300 dimensional embeddings.

A word can be tokenised into multiple subwords by both LST and LIT methods.
For the purpose of  composing the embedding of a word from the embeddings of its subwords, \newcite{zhu-etal-naacl2019}
compared vector addition, elementiwse multiplication and self-attention-based composition~\cite{Lin:ICLR:2017}.
They found vector addition to outperform other composition methods across languages and tasks.
On the other hand, prior work on sentence embedding have shown that a weighted-average of word embeddings to produce simple yet surprisingly accurate sentence embeddings~\cite{Arora:ICLR:2017,ethayarajh-2018-unsupervised}.
Inspired by these prior findings, we propose and compare three methods for composing a word embedding from its subword embeddings as follows:
\begin{description}
\item[unweighted:]
This is the simple unweighted vector addition that reported the best performance in \newcite{zhu-etal-naacl2019}.
\item[weighted:]
We use the Smoothed Inverse Frequency (SIF)~\cite{Arora:ICLR:2017}, where a word embedding $\vec{w}$ is computed as the sum of its constituent set of subwords, $\cS(w)$, weighted by their inverse unigram probabilities, $p(x)$, for subwords $x \in \cS(w)$ as given by \eqref{eq:weighted}.
\begin{align}
 \label{eq:weighted}
 \vec{w} = \sum_{x \in \cS(w)} \frac{a}{a + p(w)} \vec{x}
\end{align}
Here, the smoothing parameter $a$ is set to $0.001$ following~\newcite{Arora:ICLR:2017}. 
\item[weighted + PC removal:]
After creating word embeddings using \eqref{eq:weighted}, we substract the first Principal Component (PC) as suggested by \newcite{Arora:ICLR:2017} to remove information that is common to all words, thereby emphasising the relative semantic differences among words.
\end{description}

\subsection{Datasets and Evaluation Measures}

% Two datasets, Pearson/Spearman and the harmonic mean

To evaluate the word embeddings created using different tokenisation and composition methods described above, 
we use the datasets created for \emph{en}, \emph{de}, \emph{fr}, \emph{es}, \emph{it} and \emph{fa} in  SemEval 2017 Task 2~\cite{camachocollados-EtAl:2017:SemEval} monolingual word similarity evaluation task.
For \emph{ja} we used the dataset created by~\newcite{kodaira-etal-2016-controlled} via crowd sourcing for evaluating lexical simplification rules, which covers word-pairs categorised into different PoS categories.
For \emph{th}, we used Thai SimLex-999 dataset created by \newcite{1904.04307}. They first translated the word-pairs in the English SimLex-999~\cite{SimLex} and then asked 16 annotators, who are native Thai speakers, score the word-pairs for similarity, following the guidelines of SimLex-999~\cite{SimLex}.
For \emph{tr}, we used the AnlamVer dataset~\cite{ercan-yildiz-2018-anlamver} contains relatedness and similarity ratings for 500 Turkish word-pairs, annotated by 12 human annotators.
Following the official evaluation measure used in SemEval 2017 Task 2, on all datasets we report the harmonic mean of the Spearman and the Pearson correlation coefficients, computed between human similarity ratings and cosine similarities between the words computed using their subword-composed embeddings.

\section{Results}
\label{sec:results}

% semantic similarity
% speed (scalability with the corpus size measured in the number of sentences)

\begin{table*}[t!]
\centering
\begin{tabular}{c c c c  c c c c c c c} \toprule
 Composition & model & ~~N    & de    & en    & es    & fa    & it    & ja    & th & tr  \\  \midrule
  \parbox[t]{1mm}{\multirow{11}{*}{\rotatebox[origin=c]{90}{unweighted}}}  & \cellcolor{LightCyan} LST    \cellcolor{LightCyan}&  \cellcolor{LightCyan} 50K  & \cellcolor{LightCyan} 34.95 & \cellcolor{LightCyan} 52.89 & \cellcolor{LightCyan} 55.15 & \cellcolor{LightCyan} 50.01 &  \cellcolor{LightCyan}49.62 & \cellcolor{LightCyan} 9.66    & \cellcolor{LightCyan} 55.75 & \cellcolor{LightCyan} 26.81 \\
          &    \cellcolor{LightCyan}  & \cellcolor{LightCyan} 100K & \cellcolor{LightCyan} 48.35 & \cellcolor{LightCyan} 58.90  & \cellcolor{LightCyan}61.41 & \cellcolor{LightCyan}50.15 & \cellcolor{LightCyan}61.01 & \cellcolor{LightCyan}13.04   & \cellcolor{LightCyan} 55.57 & \cellcolor{LightCyan} 27.06\\	
          &      \cellcolor{LightCyan} & \cellcolor{LightCyan} 1M   & \cellcolor{LightCyan}50.07 & \cellcolor{LightCyan}63.29 & \cellcolor{LightCyan}64.78 & \cellcolor{LightCyan}50.42 & \cellcolor{LightCyan}62.42 & \cellcolor{LightCyan}14.37   & \cellcolor{LightCyan} 52.88 & \cellcolor{LightCyan} 20.27\\
           &   \cellcolor{LightCyan}    & \cellcolor{LightCyan}10M  & \cellcolor{LightCyan}54.10  & \cellcolor{LightCyan}63.80  & \cellcolor{LightCyan}66.20  & \cellcolor{LightCyan}50.57 & \cellcolor{LightCyan}64.88 & \cellcolor{LightCyan}14.85   & \cellcolor{LightCyan} 54.90 & \cellcolor{LightCyan} 20.52 \\
           
           & \cellcolor{LightYellow} LM    & \cellcolor{LightYellow}20K  & \cellcolor{LightYellow}52.17 & \cellcolor{LightYellow}55.61 & \cellcolor{LightYellow}53.78 & \cellcolor{LightYellow}58.59 & \cellcolor{LightYellow}52.89 & \cellcolor{LightYellow}21.11 & \cellcolor{LightYellow} 35.52 & \cellcolor{LightYellow}  35.34\\
        &  \cellcolor{LightYellow}     & \cellcolor{LightYellow}50K  & \cellcolor{LightYellow}60.66 & \cellcolor{LightYellow}65.05 & \cellcolor{LightYellow}60.72 & \cellcolor{LightYellow}59.48 & \cellcolor{LightYellow}60.91 & \cellcolor{LightYellow}22.77  & \cellcolor{LightYellow} 31.36& \cellcolor{LightYellow} 32.51\\
       &   \cellcolor{LightYellow}    & \cellcolor{LightYellow}100K & \cellcolor{LightYellow}63.38 & \cellcolor{LightYellow}66.55 & \cellcolor{LightYellow}65.46 & \cellcolor{LightYellow}59.19 & \cellcolor{LightYellow}62.29 & \cellcolor{LightYellow}18.00    & \cellcolor{LightYellow} 32.37 & \cellcolor{LightYellow}  36.47\\
            &  \cellcolor{LightYellow}     & \cellcolor{LightYellow}1M   & \cellcolor{LightYellow}59.46 & \cellcolor{LightYellow}63.06 &\cellcolor{LightYellow} 64.85 & \cellcolor{LightYellow}58.62 & \cellcolor{LightYellow}62.71 & \cellcolor{LightYellow}5.33   & \cellcolor{LightYellow} 35.59 & \cellcolor{LightYellow} 33.51 \\
            
            & \cellcolor{LightPink}BPE   & \cellcolor{LightPink} 20K  & \cellcolor{LightPink}49.41 & \cellcolor{LightPink}51.47 & \cellcolor{LightPink}52.86 & \cellcolor{LightPink}55.09 & \cellcolor{LightPink}55.73 & \cellcolor{LightPink}18.82   &  \cellcolor{LightPink} 49.56 &  \cellcolor{LightPink} 35.34 \\
            & \cellcolor{LightPink}      & \cellcolor{LightPink}50K  & \cellcolor{LightPink}58.33 & \cellcolor{LightPink}63.63 & \cellcolor{LightPink}59.54 & \cellcolor{LightPink}59.33 & \cellcolor{LightPink}60.56 & \cellcolor{LightPink}13.76   &  \cellcolor{LightPink} 53.98 & \cellcolor{LightPink} 37.00\\
            &   \cellcolor{LightPink}    & \cellcolor{LightPink}100K & \cellcolor{LightPink}61.33 & \cellcolor{LightPink}63.98 & \cellcolor{LightPink}62.79 & \cellcolor{LightPink}58.60  & \cellcolor{LightPink}63.26 & \cellcolor{LightPink}14.20   & \cellcolor{LightPink} 52.86 & \cellcolor{LightPink} 31.96 \\ \midrule
            
\parbox[t]{1mm}{\multirow{11}{*}{\rotatebox[origin=c]{90}{weighted}}}   & \cellcolor{LightCyan} LST   & \cellcolor{LightCyan} 50K  & \cellcolor{LightCyan} 34.80  & \cellcolor{LightCyan} 53.34 & \cellcolor{LightCyan} 55.90  & \cellcolor{LightCyan} 50.18 & \cellcolor{LightCyan} 49.70  & \cellcolor{LightCyan} 13.20    & \cellcolor{LightCyan} 57.01 & \cellcolor{LightCyan}  26.81\\
            &   \cellcolor{LightCyan}     & \cellcolor{LightCyan} 100K & \cellcolor{LightCyan} 48.08 & \cellcolor{LightCyan} 59.17 & \cellcolor{LightCyan} 62.80  & \cellcolor{LightCyan} 50.58 & \cellcolor{LightCyan} 61.74 & \cellcolor{LightCyan} 13.84 & \cellcolor{LightCyan}  56.84 & \cellcolor{LightCyan}  27.06 \\
            &   \cellcolor{LightCyan}     & \cellcolor{LightCyan} 1M   & \cellcolor{LightCyan} 50.08 & \cellcolor{LightCyan} 63.04 & \cellcolor{LightCyan} 67.37 & \cellcolor{LightCyan} 50.41 & \cellcolor{LightCyan} 63.55 & \cellcolor{LightCyan} 21.46  & \cellcolor{LightCyan} 54.07  & \cellcolor{LightCyan}  20.27 \\
            &    \cellcolor{LightCyan}    & \cellcolor{LightCyan} 10M  & \cellcolor{LightCyan} 54.01 & \cellcolor{LightCyan} 63.40  & \cellcolor{LightCyan} 68.55 & \cellcolor{LightCyan} 50.57 & \cellcolor{LightCyan} 65.74 & \cellcolor{LightCyan} 22.04   & \cellcolor{LightCyan}  56.06 & \cellcolor{LightCyan} 20.52 \\
            
            & \cellcolor{LightYellow}LM    & \cellcolor{LightYellow}20K  & \cellcolor{LightYellow}51.80  & \cellcolor{LightYellow}56.99 & \cellcolor{LightYellow}54.08 & \cellcolor{LightYellow}58.05 & \cellcolor{LightYellow}53.35 & \cellcolor{LightYellow}23.72   &  \cellcolor{LightYellow} 61.43&  \cellcolor{LightYellow} 32.73\\
            &  \cellcolor{LightYellow}     & \cellcolor{LightYellow}50K  & \cellcolor{LightYellow}60.4  & \cellcolor{LightYellow}65.34 & \cellcolor{LightYellow}61.58 & \cellcolor{LightYellow}58.49 & \cellcolor{LightYellow}60.95 & \cellcolor{LightYellow}27.00      &  \cellcolor{LightYellow} 61.03 &  \cellcolor{LightYellow} 33.19\\
            & \cellcolor{LightYellow}      & \cellcolor{LightYellow}100K & \cellcolor{LightYellow}63.93 & \cellcolor{LightYellow}66.48 & \cellcolor{LightYellow}66.57 & \cellcolor{LightYellow}58.25 & \cellcolor{LightYellow}62.75 & \cellcolor{LightYellow}26.80   &  \cellcolor{LightYellow} 58.18&  \cellcolor{LightYellow} 34.27 \\
            &   \cellcolor{LightYellow}    & \cellcolor{LightYellow}1M   & \cellcolor{LightYellow}59.85 & \cellcolor{LightYellow}62.62 & \cellcolor{LightYellow}66.50  & \cellcolor{LightYellow}57.83 & \cellcolor{LightYellow}64.23 & \cellcolor{LightYellow}20.77  &  \cellcolor{LightYellow} 36.15 &  \cellcolor{LightYellow} 32.28\\
            
            & \cellcolor{LightPink}BPE   & \cellcolor{LightPink}20K  & \cellcolor{LightPink}51.84 & \cellcolor{LightPink}51.07 & \cellcolor{LightPink}53.56 & \cellcolor{LightPink}54.52 & \cellcolor{LightPink}55.69 & \cellcolor{LightPink}22.95   & \cellcolor{LightPink} 51.49 & \cellcolor{LightPink} 32.73\\
            &  \cellcolor{LightPink}     & \cellcolor{LightPink}50K  & \cellcolor{LightPink}58.8  & \cellcolor{LightPink}63.69 & \cellcolor{LightPink}60.30  & \cellcolor{LightPink}58.65 & \cellcolor{LightPink}61.10  & \cellcolor{LightPink}22.25   & \cellcolor{LightPink} 55.78 & \cellcolor{LightPink} 35.99\\ 
            &  \cellcolor{LightPink}     & \cellcolor{LightPink}100K & \cellcolor{LightPink}61.67 & \cellcolor{LightPink}64.24 & \cellcolor{LightPink}63.91 & \cellcolor{LightPink}58.34 & \cellcolor{LightPink}63.22 & \cellcolor{LightPink}21.90    & \cellcolor{LightPink} 53.80& \cellcolor{LightPink} 29.57\\ \midrule
            
\parbox[t]{1mm}{\multirow{11}{*}{\rotatebox[origin=c]{90}{weighted + PC removal}}}   & \cellcolor{LightCyan}LST   & \cellcolor{LightCyan}50K  & \cellcolor{LightCyan}38.90  & \cellcolor{LightCyan}55.00    & \cellcolor{LightCyan}59.73 & \cellcolor{LightCyan}54.52 & \cellcolor{LightCyan}53.52 & \cellcolor{LightCyan}19.16 &  \cellcolor{LightCyan} 63.58 &  \cellcolor{LightCyan}  27.09\\
            &  \cellcolor{LightCyan}     & \cellcolor{LightCyan}100K & \cellcolor{LightCyan}51.82 & \cellcolor{LightCyan}61.84 & \cellcolor{LightCyan}67.43 & \cellcolor{LightCyan}59.23 & \cellcolor{LightCyan}65.34 & \cellcolor{LightCyan}23.29  &  \cellcolor{LightCyan} \textbf{64.69} &  \cellcolor{LightCyan} 28.73\\
            &  \cellcolor{LightCyan}     & \cellcolor{LightCyan}1M   & \cellcolor{LightCyan}63.12 & \cellcolor{LightCyan}71.39 & \cellcolor{LightCyan}\textbf{75.41} & \cellcolor{LightCyan}60.92 & \cellcolor{LightCyan}70.53 & \cellcolor{LightCyan}\textbf{30.98}  & \cellcolor{LightCyan} 63.81 &  \cellcolor{LightCyan} 29.31\\
            &   \cellcolor{LightCyan}    & \cellcolor{LightCyan}10M  & \cellcolor{LightCyan}65.61 & \cellcolor{LightCyan}\textbf{71.49} & \cellcolor{LightCyan}74.81 & \cellcolor{LightCyan}60.01 & \cellcolor{LightCyan}\textbf{70.85} & \cellcolor{LightCyan}30.87  &  \cellcolor{LightCyan} 64.85 &  \cellcolor{LightCyan} 28.95\\
            
            & \cellcolor{LightYellow}LM    & \cellcolor{LightYellow}20K  & \cellcolor{LightYellow}53.79 & \cellcolor{LightYellow}57.29 & \cellcolor{LightYellow}57.83 & \cellcolor{LightYellow}59.84 & \cellcolor{LightYellow}54.30  & \cellcolor{LightYellow}25.68  & \cellcolor{LightYellow} 62.49 & \cellcolor{LightYellow} 37.18 \\
            & \cellcolor{LightYellow}      & \cellcolor{LightYellow}50K  & \cellcolor{LightYellow}62.45 & \cellcolor{LightYellow}66.69 & \cellcolor{LightYellow}65.45 & \cellcolor{LightYellow}62.72 & \cellcolor{LightYellow}63.05 & \cellcolor{LightYellow}28.97  & \cellcolor{LightYellow} 61.92 & \cellcolor{LightYellow} 38.74 \\
            & \cellcolor{LightYellow}      & \cellcolor{LightYellow}100K & \cellcolor{LightYellow}64.56 & \cellcolor{LightYellow}67.58 & \cellcolor{LightYellow}71.38 & \cellcolor{LightYellow}64.20  & \cellcolor{LightYellow}65.63 & \cellcolor{LightYellow}29.39   & \cellcolor{LightYellow} 59.33 & \cellcolor{LightYellow} 36.68\\
            &  \cellcolor{LightYellow}     & \cellcolor{LightYellow}1M   & \cellcolor{LightYellow}\textbf{68.14} & \cellcolor{LightYellow}68.23 & \cellcolor{LightYellow}74.35 & \cellcolor{LightYellow}64.26 & \cellcolor{LightYellow}70.16 & \cellcolor{LightYellow}22.29   & \cellcolor{LightYellow} 38.47 & \cellcolor{LightYellow} 32.68\\
            
            & \cellcolor{LightPink}BPE   & \cellcolor{LightPink}20K  & \cellcolor{LightPink}53.03 & \cellcolor{LightPink}52.56 & \cellcolor{LightPink}56.36 & \cellcolor{LightPink}56.86 & \cellcolor{LightPink}55.66 & \cellcolor{LightPink}23.89   & \cellcolor{LightPink} 52.04 & \cellcolor{LightPink} 37.18\\
            &   \cellcolor{LightPink}    & \cellcolor{LightPink}50K  & \cellcolor{LightPink}60.17 & \cellcolor{LightPink}64.28 & \cellcolor{LightPink}64.24 & \cellcolor{LightPink}63.19 & \cellcolor{LightPink}62.76 & \cellcolor{LightPink}21.93   & \cellcolor{LightPink} 55.92 &  \cellcolor{LightPink} \textbf{41.22}\\
            & \cellcolor{LightPink}      & \cellcolor{LightPink}100K & \cellcolor{LightPink}62.60  & \cellcolor{LightPink}65.17 & \cellcolor{LightPink}68.75 & \cellcolor{LightPink}\textbf{65.40}  & \cellcolor{LightPink}65.76 & \cellcolor{LightPink}21.80   &  \cellcolor{LightPink} 53.66 &  \cellcolor{LightPink} 28.51 \\ \bottomrule
\end{tabular}
\caption{Harmonic mean of the Spearman and Pearson correlation coefficients computed between the predicted cosine similarity scores using word embeddings and human similarity ratings for different languages. Best result for each language is bolded.}
\label{tbl:corr}
\end{table*}

\begin{table}[t]
\centering
\small
\begin{tabular}{p{5mm} c c c c c c}\toprule
Method & N & adjective & adverb & noun & verb & average \\ \midrule
LST & 50K & 7.50 & 22.80 & 20.04 & 26.29 & 19.16 \\
 & 100K & 6.99 & 28.19 & 25.35 & 32.61 & 23.28 \\
 & 1M & 24.85 & 35.67 & 27.06 & 36.34 & 30.98 \\
 & 10M & 24.63 & 35.66 & 26.70 & 36.45 & 30.86 \\ \midrule
LM & 20K & 30.66 & 21.79 & 19.85 & 30.42 & 25.68 \\
 & 50K & 32.32 & 23.97 & 24.54 & 35.05 & 28.97 \\
 & 100K & 33.33 & 26.64 & 23.87 & 33.70 & 29.38 \\
 & 1M & 24.20 & 22.64 & 12.30 & 29.99 & 22.28 \\ \midrule
BPE & 20K & 24.13 & 22.98 & 18.40 & 30.01 & 23.88 \\
 & 50K & 22.87 & 19.54 & 16.66 & 28.64 & 21.93 \\
 & 100K & 22.95 & 19.70 & 16.12 & 28.42 & 21.80 \\
\bottomrule
\end{tabular}
\caption{Harmonic mean of the Spearman and Pearson correlation coefficients computed between the predicted cosine similarity scores using word embeddings computed using the SIF method and human similarity ratings for Japanese word-pairs. Results are shown separately where both words in a word-pair belongs to a particular POS category. The final column shows the arithmetic mean over the four POS categories adjectives, adverbs, nouns and verbs.}
\label{tbl:POS}
\end{table}

Performances of different tokenisation methods and composition methods across languages are summarised in \autoref{tbl:corr}.
Among the composition methods, we see that \textbf{weighted+PC removal} (SIF) consistently outperforms both \textbf{unweighted} and \textbf{weighted} for all languages.
To the best of our knowledge, SIF has not been used before for creating word embeddings from subword embeddings.
\newcite{ethayarajh-2018-unsupervised} showed that by modifying the random walk model proposed by \newcite{Arora:TACL:2016} such that the probability of generating a word given its discourse is proportional not with the inner-product between embeddings, but with their angular distance, vector length confounding effects in SIF can be rectified to create more accurate sentence embeddings.
Given such developments, an interesting future research direction would be to apply sentence embedding methods to learn better word embeddings given a subtokenisation.

From \autoref{tbl:corr}, we see that the best performances are reported by LST for all languages except for \emph{de} and \emph{fa} where respectively LM and BPE are the best.
Interestingly, for smaller vocabulary sizes (50K, 100K), we see that LM and BPE outperform LST in each language.
Given that LIT methods have been popularly used in NMT, where decoder vocabularies are typically less than 100K, it is encouraging to see that this benefit is transferrable to other NLP tasks such as semantic similarity prediction.

In \emph{de} and \emph{fa} where morphological agglutination and partial usage of fusional features are common (e.g. in the case system), we see that LIT methods such as BPE and LM outperform LST.
For example, spaCy \emph{de} tokeniser does not split compounds such as \emph{selbstfahrendes} (\emph{selbst} = self, \emph{fahren} = drive, \emph{des} = a conjugative suffix), while LM with vocabulary size 100K correctly splits it into \emph{selbst/fahren/des}. 
%We further investigated the effect of PoS on the \emph{ja} dataset and found that for highly inflected Japanese words such as adjectives and verbs,
%LM significantly outperformed LST (MeCab), but due to the large number of non-inflected nouns in the dataset the overall score for LST was better than LM. 
%Moreover, for \emph{ja}, LM consistently outperformed BPE across all vocabulary sizes and PoS categories. 
For \emph{th}, we see that LST trained with vocabulary sizes of 50K and 100K perform better than other settings.
On the other hand, LM trained with a vocabulary size of 20K performs comparably to the best LST settings for \emph{th}.
Similar to \emph{th}, for \emph{tr} we see that the LIT methods trained with vocabularies of sizes 50K and 100K perform better than other settings.
In particular, for \emph{tr} LIT consistently outperforms LST. 
This can be explained by the fact that \emph{tr} being a highly inflectional language with a derivational morphology.
This result reinforces the observation that subword tokenisation via LIT methods is particularly effective for strong inflective languages such as \emph{ja} and \emph{tr}, when creating word embeddings.

Given the language independent nature of LIT methods, an interesting question is whether it would be beneficial to train a single LIT tokeniser for a group of languages.
To address this question,iIn a preliminary study, we mixed all corpora in \autoref{tbl:corp} to create a single multilingual corpus and trained LM and BPE on it.
However, the tokeniser models obtained by this approach were poor, which suggests that LIT methods must be trained on monolingual corpora.
This could be due to the disproportions of the sizes of the corpora available for different languages, which bias the subtoken statistics for some languages than the others.
Careful data sampling would be needed to create balanced text corpora for learning universal LIT models.
Investigating methods for learning universal LIT models is beyond the scope of the current paper and would be an interesting future research direction.

\subsection{Effect of Part-of-Speech}

To further study the effective of tokenisation for different POS categories, we use the Japanese word similarity dataset created by \newcite{kodaira-etal-2016-controlled}. This dataset classifies word-pairs according to POS category of the two words being compared. In particular, both words in a word-pair belong to the same POS category, which makes it an ideal candidate for studying the effect of tokenisation on different POS categories.
As observed in \autoref{tbl:corr}, among the different composition methods, SIF method reported the best results across languages. 
Therefore, we use SIF for creating word embeddings from subword embeddings in this experiment.
Specifically, we use the GloVe embeddings for Japanese subtokens/tokens obtained by a particular tokenisation method and use SIF to create the word embeddings for each word in word-pairs in the Japanese semantic similarity dataset. 
The similarity between two words is computed by the cosine of the angle between the corresponding word embeddings.
Next, we measure the Spearman and Pearson correlation between the predicted similarity scores and the human ratings for each POS category and report the harmonic mean between the Spearman and Pearson correlation coefficients as done in the previous experiment.
Arithmetic mean (average) over the four POS categories -- adjectives, adverbs, nouns and verbs, are reported in \autoref{tbl:POS}.

% adjectives are highly inflected and small vocab sizes with LST perform worse. Increasing the vocab helps but there is a significant gain by LM even with smaller vocabs for adjectives.
% Increasing vocab size shows a continuous improvement for LST. But with LM it increases from 20K to 50K but then steadily drops. This is because of the over splitting to smaller subtokens and computing word embeddings from such smaller subtokens that possibly do not retain sufficient semantics is problematic. This issue is prominent especially for nouns.
% Similar trend can be seen for BPE as well, which implies this is an issue for LIT methods. When larger vocabularies are used one must be careful to appropriately fine tune the vocabulary size, which is a hyperparameter.

From \autoref{tbl:POS}, we see that the performance of LST with smaller vocabularies such as 50K or 100K tokens for adjectives is poor.
Compared to other POS categories, adjectives are highly inflected in Japanese and depend on the tense of the sentence.
Therefore, a smaller vocabulary might not be sufficient to cover all the variants of adjectives.
On the other hand, LIT methods such as LM significantly outperforms LST even with a smaller vocabulary size of 20K subtokens.
This result reinforces the observation we made in \autoref{tbl:corr} that LIT methods are attractive for obtaining good performance with smaller vocabulary sizes.
Increasing the size of the vocabulary results in a steady improvement in performance for LST. However, the same cannot be said about LIT.
For example, the performance of LM increases when the size of the vocabulary is increased from 20K to 50K but drops when it is increased beyond 50K.
This issue is particularly sever for nouns.
Similar trends can be observed with BPE as well.

Larger vocabularies contain many smaller subtokens and the probability of a given text getting over-tokenised into many smaller tokens increases with the size of the vocabulary for LIT. 
Creating the embedding for a word using embeddings for its subtokens becomes difficult when the word is split into many subtokens, some of which might be too small to retain the semantics of the original word.
Recall that SIF method creates word embeddings as the weighted-average of the subtoken embeddings, ignoring the position of the subtoken in the word.
Incorporating character-level embeddings via LSTMs has shown to improve performance for named entity recognition tasks~\cite{zhai-etal-2018-comparing}.
Therefore, applying more sophisticated supervised composition methods such as  a recurrent neural network might help to create word embeddings from subtoken embeddings under such situations.
We defer this line of investigation for future work.
We conclude here that the size of the vocabulary is a hyperparameter of LIT methods that must be carefully set considering the performance of the target task.

\subsection{Nearest Neighbour Analysis}
\label{sec:NN}

Given that some subtokens correspond to character $n$-grams representing morphology such as inflections, it remains an interesting qualitative analysis to study whether such information is encoded in the learnt subword embeddings. 
We select prefixes or suffixes that have known inflectional roles and compute the cosine similarity between each prefix/suffix and all other tokens in the vocabulary using the unweighted embedding method to find the nearest neighbours in the embedding space.
Specifically, we conduct this nearest neighbour analysis for the three languages: English, Japanese and Turkish.
English is selected as a language that uses the space character to denote word boundaries, Turkish and Japanese are selected as agglutinative languages, whereas word boundaries are not marked by the space character in Japanese.
We use the LIT models obtained using LM with vocabulary sizes 100K, 50K and 50K respectively for English, Japanese and Turkish for finding the nearest neighbours using subword embeddings.

\autoref{tbl:en} shows the nearest neighbours for  the suffixes \emph{ed} and \emph{ing}, which often inflects verb tense in English.
We use an underscore to denote a token boundary corresponding to the space character. 
From \autoref{tbl:en}, we see that verbs that are frequently inflected using those suffixes are ranked at the top as the nearest neighbours, indicating that the relationship between inflective suffixes and verbs is preserved during LIT. 

\autoref{tbl:ja} shows the nearest neighbours for the Japanese verb ending form \emph{masu}. 
We see that various inflections of \emph{masu} are listed as the top nearest neighbours such as its past tense (\emph{mashita}), negation (\emph{masen}) and the volitional form (\emph{mashou}).
We also see that other frequent sentence ending forms such as \emph{desu} and \emph{kudasai} are also listed as nearest neighbours.
Similar trends have been reported with distributional word-level embeddings, where both semantically similar as well as related/associated words are often found as the nearest neighbours for a given word when the cosine similarity between word embeddings is used as the neighbourhood criterion~\cite{SimLex,weeds-EtAl:2014:Coling}.
\autoref{tbl:tr} shows the nearest neighbours for the Turkish suffixes \emph{iyor} and \emph{miyor}, which respectively denote the present tense and its negation.
Likewise in English and Japanese results, we see related words are listed as the nearest neighbours for those suffixes.
However, the nearest neighbours retrieved in the case of Turkish are more noisier compared to that for English and Japanese.
We believe this is due to the comparatively smaller corpora used for Turkish.

\begin{table}[t]
\centering
%\small
\begin{tabular}{l l}\toprule
\emph{ing} & \emph{ed} \\\midrule
ed (0.610188) & ing (0.610188) \\
\_utiliz (0.416099) &\_aggravat (0.3683) \\
\_consolidat (0.4143) & \_dispos (0.3682) \\
\_thereby (0.4138) & \_encas (0.3670) \\
\_manipulat (0.4125) &\_accentuat (0.3666) \\
\_incorporat (0.4029) &\_clipp (0.3634) \\
\_facilitat (0.3980) &\_precipitat (0.3600) \\
\_expell (0.3937) &\_produc (0.3580) \\
\_involves (0.3916) & \_exacerbat (0.3576) \\
\_dedicat (0.3895) & \_rechristen (0.3543) \\
\_without (0.3841) &\_supplant (0.3498) \\
\bottomrule
\end{tabular}
\caption{Nearest neighbours and their cosine similarity scores (indicated within brackets) for the two English suffixes \emph{ing} and \emph{ed}.}
\label{tbl:en}
\end{table}

\begin{table*}[t]
\small
\centering
\begin{tabular}{l l}\toprule
Analogy & Top candidates \\ \midrule
\textbf{\_improving} - \textbf{ing} + \textbf{ed} & \_improved (0.5817), \_improve (0.5791), \_prioritize (0.4582), \_improvement (0.4526) \\
\textbf{\_posterior} - \textbf{\_prior} + \textbf{\_pre} & \_anterior (0.6705), \_dorsal (0.5405), \_medial (0.5341), \_ventral (0.5266) \\
\textbf{\_export} - \textbf{\_ex} + \textbf{\_im} & \_exports (0.4191), \_markets (0.3998), \_exporting (0.3906), \_importation (0.3850) \\ \midrule
\textbf{\jp{しない}} - \textbf{\jp{する}} + \textbf{\jp{ます}} &  \jp{ません} (0.6492), \jp{ました} (0.6344), \jp{です} (0.5570), \jp{絶対に} (0.5551)\\
\textbf{\jp{こんな}} - \textbf{\jp{ここ}} + \textbf{\jp{そこ}} & \jp{そんな} (0.6054), \jp{どんな} (0.5686), \jp{そういう} (0.5634), \jp{なんて} (0.5627) \\
\textbf{\jp{直前}} - \textbf{\jp{前}} + \textbf{\jp{後}} & \jp{直後} (0.5424), \jp{直後に} (0.5046), \jp{後に} (0.4738), \jp{直前に} (0.4677) \\
 \midrule
\textbf{meyecek} - \textbf{ecek} + \textbf{iyor} & \_miyor (0.5093), \_miyordu (0.4835), \_iyordu (0.4828), \_mekteydi (0.4459)\\
\textbf{\_bunu} - \textbf{nu} + \textbf{na} & \_buna (0.5207), \_rağmen (0.4771), \_fakat (0.4610), \_ama (0.4530) \\
\textbf{\_gitmek} - \textbf{mek} + \textbf{ti} & \_gönderilmiş (0.4135), \_gitti (0.4078), ten (0.3550), \_yola (0.3424) \\
\bottomrule
\end{tabular}
\caption{Top candidates for the analogies ranked according to their cosine similarity (shown within brackets) with the target vector for English, Japanese and Turkish.}
\label{tbl:ana}
\end{table*}

\begin{table}[t]
\centering
\small
\begin{tabular}{l p{8mm} c}\toprule
\jp{ます} (\emph{masu}) & similarity & info \\\midrule
\jp{ました} (\emph{mashita}) & 0.8273 & conjugation of \emph{masu} \\
\jp{ません} (\emph{masen}) &  0.6522 & conjugation of \emph{masu} \\
\jp{ましょう} (\emph{mashou})& 0.6400 & conjugation of \emph{masu} \\
\jp{ください} (\emph{kudasai})& 0.5846 & imperative verb for \emph{please} \\
\jp{です} (\emph{desu}) & 0.5819 & sentence ending\\
\jp{なさい} (\emph{nasai}) & 0.5514 & imperative verb for \emph{do} \\
\bottomrule
\end{tabular}
\caption{Nearest neighbours and their cosine similarity scores for the Japanese verb \emph{masu}. }
\label{tbl:ja}
\end{table}

\begin{table}[t]
\centering
\begin{tabular}{l  l}\toprule
\emph{iyor} & \emph{miyor} \\\midrule
iyordu  (0.7491) &  miyordu (0.6765) \\
miyor (0.5655) & iyor (0.5655) \\
ecektir  (0.5349) & miyorsa (0.5370) \\
eceğini (0.5097) & \_destekle (0.5130) \\
eceği (0.5036) & \_gel (0.5033) \\
\_geçir (0.5005) & ebiliyordu (0.5012) \\
iyorum (0.4971) & ememektedir (0.4971) \\
mektedir (0.4911) & ebiliyor (0.4961) \\
\bottomrule
\end{tabular}
\caption{Nearest neighbours and their cosine similarity scores (indicated within brackets) for, \emph{iyor} and \emph{miyor} the Turkish suffixes indicating respectively the present tense and its negation.}
\label{tbl:tr}
\end{table}

Word embedding spaces learnt by word2vec and GloVe have shown to demonstrate a surprisingly high degree of relational structure, which can be exploited to solve analogies~\cite{Allen:ICML:2019,Mikolov:NAACL:2013}. 
To test whether these relational properties exist in subtoken embedding spaces, we use the unweighted word embeddings and solve exemplar analogies as shown in \autoref{tbl:ana}.
Specifically, for an analogy ``$a$ is to $b$ as $c$ is to d'', given $a$, $b$ and $c$ we find candidates $d$ that satisfy the analogy according to the cosine similarity between the vector $\vec{b} - \vec{a} + \vec{c}$ and each of the subtoken embedding $\vec{d}$ in the vocabulary. 
We then rank the candidates $d$ in the descending order of the cosine similarity scores.
The first set of three rows in \autoref{tbl:ana} show analogies for English, whereas the second and third sets of three rows respectively show analogies for Japanese and Turkish.

For English we see that suffixes (as in the case for \emph{improve}) as well as prefixes (as in the case for \emph{export}) demonstrate a certain level of relational structure in the embedding space.
However, analogies are not always correctly preserved in the subtoken embedding spaces as seen from the example for \emph{posterior}.
Although \emph{anterior} (front of the body) and \emph{dorsal} (upper side or back of an animal or plant) are closely related to the semantics implied by the resulting vector, they are not perfect candidates.
On the other hand, the Japanese subtoken embeddings show interesting analogical structures.
For example, subtracting the embedding for the kanji character \jp{前} (\emph{mae}, meaning \emph{before}) from \jp{直前} (\emph{chokuzen}, meaning \emph{immediately before}) and adding \jp{後} (\emph{ato}, meaning \emph{after}), we can discover \jp{直後} (\emph{chokugo}, meaning \emph{immediately after}).
We see that the Turkish suffixes \emph{ecek} (indicating the future tense) and \emph{meyecek} (indicating the negated future tense) form the analogy \textbf{meyecek} - \textbf{ecek} + \textbf{iyor} with \emph{miyor}.
Overall, we see that the analogical relationships reported for word embeddings in prior work can also be seen with subword embeddings.

\subsection{Training time}

LIT methods such as BPE and LM must be first trained on an untokenised corpus to compute the vocabularies and the frequencies of the subtokens.
Larger corpora that cover various word forms are desirable for this purpose because it enables us to obtain reliable subtoken frequencies for a larger vocabulary.
However, the training time depends on the size of the vocabulary and is an important aspect to consider in practice.

\begin{figure}[t]
\centering
\includegraphics[width=8.5cm]{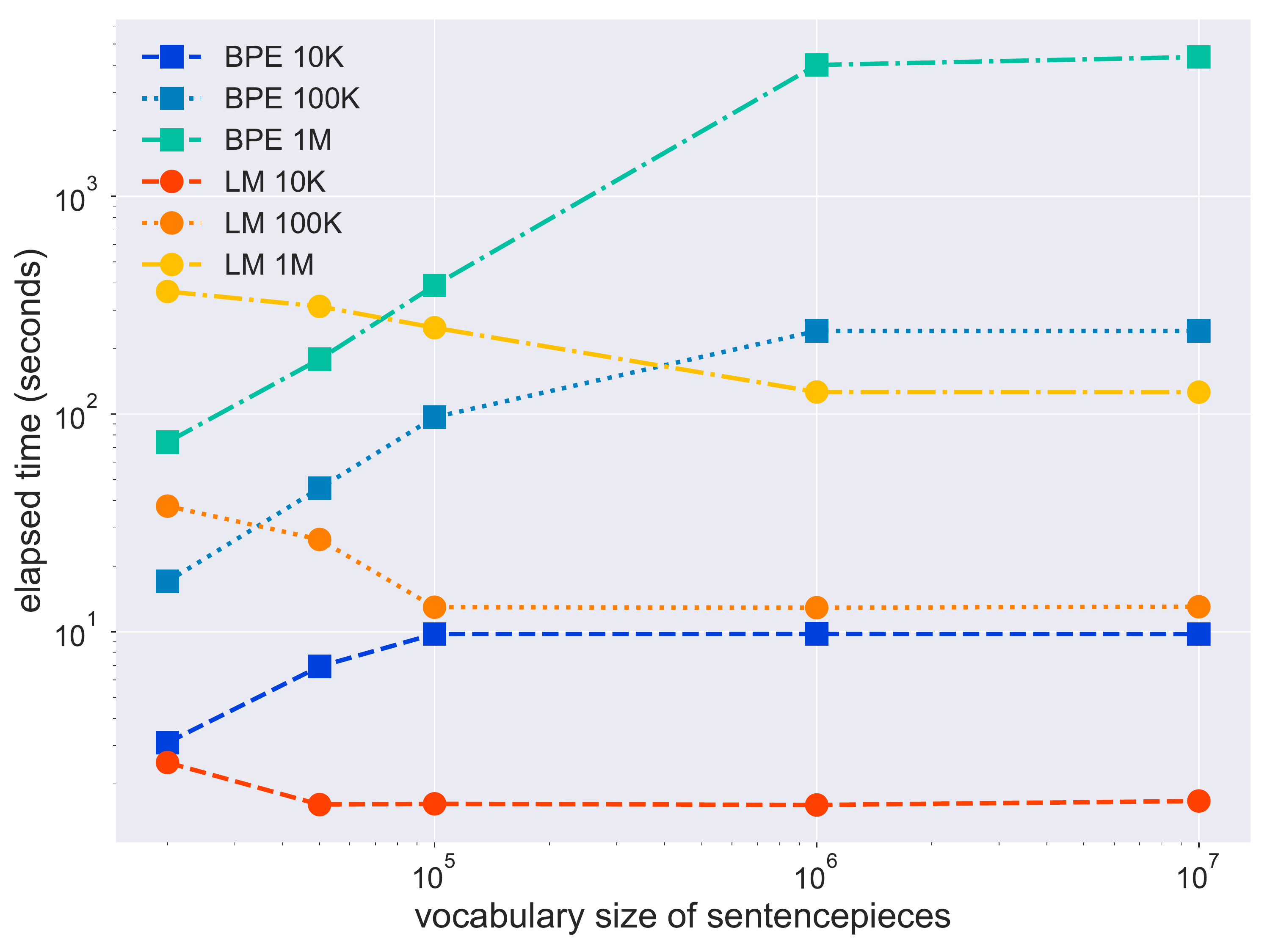}
\caption{Comparison of the training time for BPE and LM for different numbers of input sentences.}
\label{fig:speed}
\end{figure}

In \autoref{fig:speed}, to study the scalability of LIT methods, 
we compare BPE and LM in a single threaded setting on the same hardware (m5.24xlarge AWS instances) under different numbers of input sentences.
To the best of our knowledge, prior work comparing LIT methods have not studied the effect on training time for LM or BPE.
From \autoref{fig:speed}, we see that LM is significantly faster than BPE, and its training time decreases with the vocabulary size, while the opposite is true for BPE.
This is because BPE iteratively increases the vocabulary until the desired size is reached, whereas LM iteratively decreases the same.
Moreover, further speed ups for LM can be easily obtained via multi-threading because the E-step of the likelihood computation in LM is embarrassingly parallelisable.

\section{Conclusion}
\label{sec:conclusion}

We compared LST against two LIT methods (BPE and LM) for multiple languages using similarity prediction as an evaluation task.
After tokenising a text corpus, we used GloVe to learn embeddings for the subtokens. 
Next, we created word embeddings by composing the subtoken embeddings.
We used semantic similarity prediction as a evaluation task where we predict the similarity between two words by the cosine of the angle between the corresponding word embeddings.
We found that when the vocabulary size is large, LST methods consistently outperform LIT methods.
However, for smaller vocabularies (less than 100K), LIT methods outperformed LST methods, suggesting that LIT is suitable for resource poor languages or when smaller models are required.
Moreover, SIF method, which weights subword embeddings by unigram probability and subtract the first principal component vector was found to be an effective composition method for creating word embeddings from subword embeddings.
We analysed the nearest neighbours for subtokens and found that semantically and syntactically related subtokens are retrieved as the top nearest neighbours using subword embeddings.
Moreover, analogical structures, which have been previously reported for word embedding spaces, can also be found even in subword embedding spaces.

\section{Bibliographical References}
\bibliographystyle{lrec}
\bibliography{subtok}

\end{document}